\documentclass[journal]{IEEEtran}
\ifCLASSINFOpdf
\else
\usepackage[dvips]{graphicx}
\fi

\usepackage{cite}
\usepackage{booktabs}
\usepackage[pdftex]{graphicx}
\usepackage[cmex10]{amsmath}
\usepackage{multirow}
\usepackage{color}
\usepackage{mathtools}
\usepackage{array}
\usepackage{dblfloatfix}
\usepackage{subfig}
\usepackage{bm}

\usepackage{amsmath,amsfonts,bm}

\def\eqref#1{equation~\ref{#1}}

\def\1{\bm{1}}

\def\va{{\bm{a}}}

\def\vd{{\bm{d}}}
\def\ve{{\bm{e}}}

\def\vs{{\bm{s}}}
\def\vt{{\bm{t}}}
\def\vu{{\bm{u}}}
\def\vv{{\bm{v}}}

\def\vx{{\bm{x}}}

\def\vz{{\bm{z}}}

\DeclareMathAlphabet{\mathsfit}{\encodingdefault}{\sfdefault}{m}{sl}
\SetMathAlphabet{\mathsfit}{bold}{\encodingdefault}{\sfdefault}{bx}{n}

\usepackage{balance}
\usepackage{url}
\usepackage{multirow, float, booktabs, boldline, hhline}
\usepackage{cellspace}
\usepackage{color, colortbl}
\usepackage{algorithm}
\usepackage{algpseudocode}
 \usepackage{xcolor}

\usepackage{cite}

\begin{document}
\title{GSMFlow: Generation Shifts Mitigating Flow for Generalized Zero-Shot Learning}


\author{Zhi~Chen, Yadan~Luo, Sen~Wang, Jingjing~Li, Zi~Huang
\thanks{Z. Chen, Y. Luo, S. Wang and Z. Huang are with School of Information Technology \& Electrical Engineering, The University of Queensland, Brisbane, QLD 4072, Australia. (e-mails:zhi.chen@uq.net.au, lyadanluol@gmail.com, sen.wang@uq.edu.au, huang@itee.uq.edu.au).}
\thanks{J. Li is with the School of Computer
Science and Engineering, University of Electronic Science and Technology of
China (email: jjl@uestc.edu.cn).}
}

\markboth{IEEE Transactions on Multimedia, XXXX. 2020}%
{Shell \MakeLowercase{\textit{et al.}}: Bare Demo of IEEEtran.cls for Journals}

\maketitle

\begin{abstract}

Generalized Zero-Shot Learning (GZSL) aims to recognize images from both the seen and unseen classes by transferring semantic knowledge from seen to unseen classes.
It is a promising solution to take the advantage of generative models to hallucinate realistic unseen samples based on the knowledge learned from the seen classes.
However, due to the generation shifts, the synthesized samples by most existing methods may drift from the real distribution of the unseen data.
To address this issue, we propose a novel flow-based generative framework that consists of multiple conditional affine coupling layers for learning unseen data generation.
Specifically, we discover and address three potential problems that trigger the generation shifts, \textit{i.e.}, \textit{semantic inconsistency}, \textit{variance collapse}, and \textit{structure disorder}.
First, to enhance the reflection of the semantic information in the generated samples, we explicitly embed the semantic information into the transformation in each conditional affine coupling layer. 
Second, to recover the intrinsic variance of the real unseen features, we introduce a boundary sample mining strategy with entropy maximization to discover more difficult visual variants of semantic prototypes and hereby adjust the decision boundary of the classifiers.
Third, a relative positioning strategy is proposed to revise the attribute embeddings, guiding them to fully preserve the inter-class geometric structure and further avoid structure disorder in the semantic space. 
Extensive experimental results on four GZSL benchmark datasets demonstrate that GSMFlow achieves the state-of-the-art performance on GZSL. 
\end{abstract}

\IEEEpeerreviewmaketitle
\section{Introduction}
\IEEEPARstart{T}{raditional} visual recognition systems have progressed substantially with the help of a massive amount of labeled data and deep learning techniques \cite{zhang2021joint,zhang2021high,zhang2021aggregation,9616392_Dong,What_Transferred_Dong_CVPR2020}. However, it is challenging for these recognition systems to generalize to unseen classes that are unknown during training. Also, the need for the massive amount of labeled data has been a curse when training a deep recognition system.
Generally, one may curate a certain amount of data of every class for training a canonical classification model, while the need for data is drastically amplified for deep learning methods.
Data labeling is time-consuming and expensive, even if the raw data is usually plentiful. Moreover, it is unrealistic to require labeled data for every class, leading to extra attention from researchers drawn to Zero-Shot Learning (ZSL) as a solution \cite{akata2015label,xian2016latent,akata2015evaluation,yang2016zero,li2019zero}. By incorporating side information, \textit{e.g.,} class-level semantic attributes, ZSL transfers semantic-visual relationships from the seen classes to unseen classes without any visual samples. While conventional ZSL aims to recognize only unseen classes, it is infeasible to assume that we will only come across samples from unseen classes. Hence, in this paper, we consider a more realistic and challenging task, Generalized Zero-Shot Learning (GZSL), to classify over both seen and unseen classes \cite{huang2019generative,schonfeld2019generalized,xian2018feature,li2020learning}.

\begin{figure}[t]
\centering
\includegraphics[width=85mm]{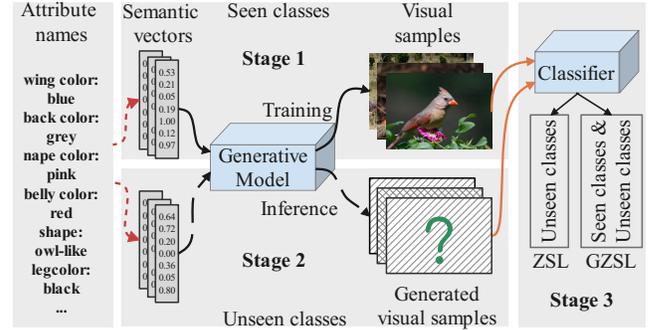}
\caption{ An illustration of generative zero-shot learning. }
\label{generative}
\end{figure}

In GZSL, there are two mainstream methods 
GZSL can be roughly categorized into embedding-based methods \cite{xian2016latent,akata2015label,akata2015evaluation,xie2020region,li2021attribute} and generative methods \cite{li2019leveraging,xian2018feature,schonfeld2019generalized,li2019alleviating,xie2021generalized,ye2022learning,su2022distinguishing}. 
Embedding-based methods usually cast the semantic and visual features into the same space, where the compatibility scores between the visual features and all classes are computed for predictions. In contrast, generative methods \cite{long2017zero,xian2018feature,schonfeld2019generalized,chen2020rethinking,shen2020invertible,li2021investigating}, as shown in Fig. \ref{generative}, cast the GZSL problem into a supervised classification task by generating synthesized visual features for unseen classes. 
Then a supervised classifier can be trained on both real seen visual features and the synthesized unseen visual features. 

\begin{figure}[t]
\centering
\includegraphics[width=85mm]{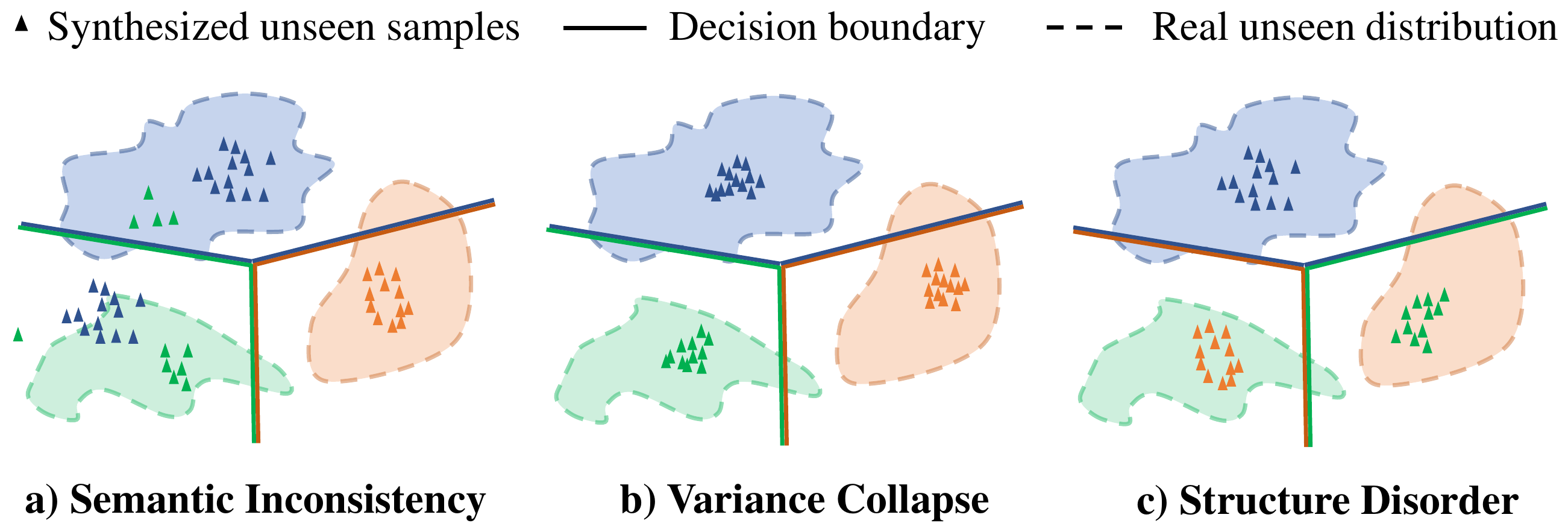}
\caption{ An illustration of the generation shifts. 
a) Implicit semantic encoding causes the semantic inconsistency when generating unseen samples. b) The synthesized samples collapse to fixed modes. c) The synthesized samples fail to fully preserve the geometric relationships in the semantic space.
}
\label{intro}
\end{figure}

In spite of the advances achieved by the generative paradigm, the synthesized visual features may not be guaranteed to cover the real distribution of unseen classes, given that the unseen features are not exposed during training.
Due to the gap between the synthesized and the real distributions of unseen features, the trained classifier tends to misclassify the unseen samples.
Therefore, it is critical to analyze the causes and mitigate the resulted generation shifts. In this paper, as illustrated in Fig. \ref{intro}, we identify three common types of generation shifts in generative zero-shot learning methods:
\begin{itemize}
    \item \textbf{Semantic inconsistency:} The state-of-the-art GZSL approach \cite{shen2020invertible} focuses on preserving the exact generation cycle consistency by a normalizing flow. It constructs a complex prior and disentangle the outputs into semantic and non-semantic vectors. Such implicit encoding may cause the generated samples to be incoherent with the given attributes and deviated from the real distributions.
    \item \textbf{Variance collapse:} The synthesized samples commonly collapse into fixed modes and fail to capture the intra-class variance of unseen samples, which can be originated from different poses, illumination and background.
     \item \textbf{Structure disorder:} It is also hard to fully preserve the geometric relationships between different attribute categories in the shared subspace. If the relative position of unseen samples changes in the visual space, the recognition performance will inevitably degenerate. 
\end{itemize}

\begin{figure*}[t]
\centering
\includegraphics[width=\textwidth]{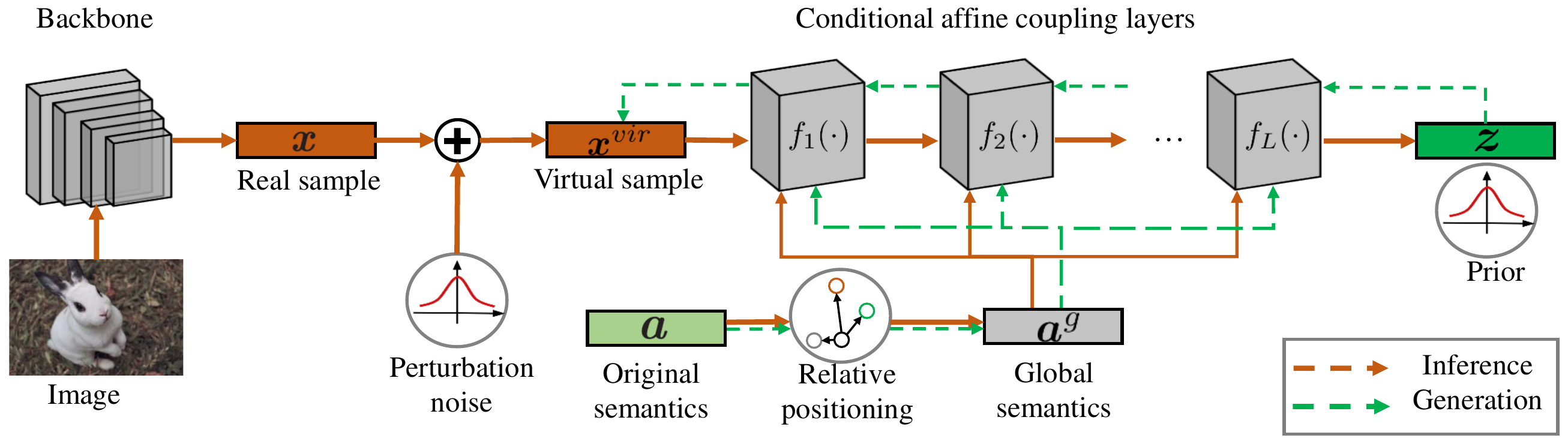}
\caption{An illustration of the proposed GSMFlow framework. The conditional generative flow is comprised of a series of conditional affine coupling layers. Particularly, the boundary samples are learned, and additional noise are injected into the visual features to complement the potential patterns. Further, the global semantics are computed with relative positioning to semantic anchors and explicitly encoded into each conditional affine coupling layer.
For inference, a latent variable $\vz$ is inferred from the visual features of an image sample $\vx$ conditioned on a global semantic vector $\va^g$. Inversely, given $\vz$ drawn from a prior distribution and a global semantic vector $\va^g$, GSMFlow can generate a visual sample accordingly.}
\label{architecture}
\end{figure*}

To address these potential generation shifts, in this paper, we propose a novel framework for generalized zero-shot learning, namely Generation Shifts Mitigating Flow (GSMFlow), as depicted in Fig. \ref{architecture}. 
In particular, to tackle the semantic inconsistency, we explicitly embed the semantic information into the transformations in each of the coupling layers during inference and generation, enforcing the unseen visual feature generation to be semantically consistent. 
Furthermore, to mitigate the variance collapse issue, a boundary sample mining technique with entropy maximization is proposed to explicitly discover the decision boundaries between classes and more visual variants that are explored with visual perturbation.
With this strategy, the generative model is exposed to more visual variants for each class, which will adjust the decision boundary for suiting the real unseen samples at the test time.
Moreover, to alleviate the semantic structure disorder, the relative positioning mechanism is adopted to correct the attribute representations by preserving the global geometrical relationship to the specific semantic anchors. 
Specifically, the visual features $\vx$ of a real sample are first extracted from a backbone network, \textit{e.g.,} ResNet101. 
In the training stage, the boundary samples that represent more extreme visual variants are learned from the visual samples in the original dataset, then we add a small amount of noise which brings more diversity. The resulted diverse and hard visual samples are fed into a flow-based generative framework.
The training process is conditioned on the manipulated global semantic vector to infer the latent factors drawn from a prior distribution. 
The learned generative flow has its inverse transformation as the generative network. 
Similarly, a global semantic vector is progressively injected into each of the inverse coupling layers to generate unseen visual features from latent factors. 
Eventually, a unified classifier is trained with the real seen features and the generated unseen features, aiming to accurately recognize both seen and unseen classes at the test time. 
To sum up, the contributions of our work are listed as follows: 
\begin{itemize}
    \item 
    A novel GSMFlow framework is proposed for GZSL, which explicitly incorporates the class-level semantic information into both the forward and the inverse transformations of the conditional generative flow. It encourages the synthesized samples to be more coherent with the respective semantic information.  
    
    \item We propose a boundary sample mining strategy to explicit learn the visual representation boundaries for each semantic vector. Moreover, a visual perturbation strategy to imitate the intrinsic variance of unseen classes in synthesized samples. By learning more boundary samples and injecting perturbation noise, we diversify the training samples to enrich the original feature space. As the generative flow is exposed to more diverse virtual samples, the synthesized samples of unseen classes can thus capture more visual potentials.
    
   
    \item To preserve the geometric relationship between different semantic vectors in the semantic space, we choose different semantic anchors to revise the representation of the attributes.
    \item Comprehensive experiments and in-depth analysis on four GZSL benchmark datasets demonstrate the state-of-the-art performance by the proposed GSMFlow framework in the GZSL tasks.
\end{itemize}

A preliminary version of GSMFlow is presented in \cite{chen2021mitigating}, which is accepted by ACM multimedia 2021. We have made extensive revision and expansion in this paper. To address the variance collapse issue, in the preliminary version, we proposed a visual perturbation strategy to imitate the intrinsic variance of unseen classes in synthesized samples.
In this paper, we further propose a boundary sample mining technique with entropy maximization that explicitly enriches the visual space to prevent variance collapse. Moreover, more detailed discussions and analysis are provided in this paper.

The rest of the paper is organised as follows. We briefly review related work in Section 2. GSMFlow is presented in Section 3, followed by the experiments and the in-depth analysis in Section 4. Lastly, Section 5 concludes the paper.

\section{Related Work}

\subsection{Traditional Zero-shot Learning}
Traditional solutions towards zero-shot learning are mostly the embedding-based methods. The pioneering method ALE \cite{akata2015label} proposes to employ embedding functions to measure the compatibility scores between a semantic embedding and a data sample. SJE \cite{akata2015evaluation} extends ALE by using structured SVM \cite{tsochantaridis2005large} and takes advantage of the structured outputs. DeViSE \cite{frome2013devise} constructs a deep visual semantic embedding model to map 4096-dimensional visual features from AlexNet \cite{krizhevsky2012imagenet} to 500 or 1000-dimensional skip-gram semantic embeddings. EZSL \cite{romera2015embarrassingly} theoretically gives the risk bound of the generalization error and connects zero-shot learning with the domain adaptation problem. More recently, SAE \cite{kodirov2017semantic} develops a cycle embedding approach with an autoencoder to reconstruct the learned semantic embeddings into a visual space. Later, the cycle architecture is also investigated in generative methods \cite{chen2020canzsl}. However, these early methods have not achieved satisfactory results on zero-shot learning. Particularly, when applying on the GZSL task, the unseen class performance is even worse. Recently, thanks to the advances of generative models, by generating missing visual samples of unseen classes, zero-shot learning can be converted into a supervised classification task.

\subsection{Generative GZSL}
A number of generative methods have been applied for GZSL, \textit{e.g.,} Generative Adversarial Nets (GANs) \cite{goodfellow2014generative}, Variational Autoencoders (VAEs) \cite{kingma2013auto}, and Alternating Back-Propagation algorithms (ABPs) \cite{han2017alternating}. f-CLSWGAN \cite{xian2018feature} presents a WGAN-based \cite{arjovsky2017wasserstein} approach to synthesize unseen visual features based on semantic information. CADA-VAE \cite{schonfeld2019generalized} proposes to stack two VAEs, each for one modality, and aligns the latent spaces. The latent space, thus, can enable information sharing among different modality sources. LisGAN \cite{li2019leveraging} is inspired by the multi-view property of images and improves f-CLSWGAN by encouraging the generated samples to approximate at least one visually representative view of samples. CANZSL \cite{chen2020canzsl} considers the cycle-consistency principle of image generation and proposes a cycle architecture by translating synthesized visual features into semantic information. ABP-ZSL \cite{zhu2019learning} adopts the rarely studied generative models ABPs to generate visual features for unseen classes. GDAN \cite{huang2019generative} incorporates a flexible metric in the model’s discriminator to measure the similarity of features from different modalities. 

The bidirectional conditional generative models enforce cycle-consistent generation and allow the generated images to truthfully reflect the conditional information \cite{zhu2017unpaired,chen2020canzsl,chen2019cycle}. Instead of encouraging cycle consistency through adding additional reverse networks, generative flows are bidirectional and cycle-consistent in nature. The generative flows are designed to infer and generate within the same network. Also, the generative flows are lightweight comparing to other methods as no auxiliary networks are needed, \textit{e.g.,} discriminator for GANs, variational encoder for VAEs. Some conditional generative flows \cite{ardizzone2018analyzing,ardizzone2019guided} are proposed to learn image generation from class-level semantic information. IZF \cite{shen2020invertible} adopts the invertible flow model \cite{ardizzone2018analyzing} for GZSL. It implicitly learns the conditional generation by inferring the semantic vectors. Such implicit encoding may cause the generated samples to be incoherent with the given attributes. Instead, we explicitly blend the semantic information into each of the coupling layers of the generative flow, learning the semantically consistent visual features. We also argue that the MMD regularization in IZF is inappropriate for GZSL, since the seen and unseen classes are coalesced. 
It enforces conditional generation by mixing the conditional vectors with the prior. However, this design weakens the conditional information through the affine coupling layers. In our framework, instead, we progressively input the conditional information in each affine coupling layer, which enhances the semantic information in the generation process.

\subsection{Entropy Maximization}
Entropy regularization in supervised learning is usually associated with confidence penalty, which is invoked for out-of-distribution detection \cite{alemi2018uncertainty}, domain generalization \cite{zhao2020maximum, wang2021learning}, and adversarial samples \cite{alemi2016deep}. By maximizing the entropy of the probability distribution, the prediction confidence for the input sample is penalized. Zhao \textit{et al.} \cite{zhao2020maximum} developed an efficient maximum-entropy regularizer to generate hard adversarial perturbations from information bottleneck principle. Alexmi \textit{et al.}  \cite{alemi2018uncertainty} investigated that variational information bottleneck can improve the classification calibration and the ability to detect out-of-distribution samples. They demonstrate that a deterministic classifier is overconfident, and a high predictive entropy signifies a network does not know how to classify the inputs. These work inspire us to consider mining boundary samples with entropy maximization to explicitly explore the hard samples in zero-shot learning. By learning more extreme visual variants that are close to negative classes, the generalization ability to unseen classes are significantly improved.



\section{Methodology}
This section begins with formulating the GZSL problem and introducing the notations. The proposed GSMFlow is outlined next, followed by the boundary sample mining, visual perturbation strategy, and the relative positioning approach to mitigate the generation shifts problem. The model training and zero-shot recognition are introduced lastly.

\subsection{Preliminaries}
Consider two datasets - a seen set $\mathcal{S}$ and an unseen set $\mathcal{U}$. The seen set $\mathcal{S}$ contains $N^s$ training samples $\vx^s \in \mathcal{X}$ and the corresponding class labels $y^s$, \textit{i.e.}, $\mathcal{S} = \{ \vx_{(i)}^{s}, y_{(i)}^{s} \}_{i=1}^{N^s}$. Similarly, $\mathcal{U} = \{ \vx_{(i)}^{u}, y_{(i)}^{u} \}_{i=1}^{N^u}$, where $N^u$ is the number of unseen samples. There are $C_s$ seen and $C_u$ unseen classes, so that $y^s \in \{1,\ldots,C_s \}$ and $y^u \in \{C_s+1,\ldots,C_s+C_u\}$. Note that the seen and unseen classes are mutually exclusive. There are $C_s+C_u$ class-level semantic vectors $\va \in \mathcal{A}$, where the class-level semantic vectors for the seen and unseen classes are $\{\va_{i}^{s}\}_{i=1}^{C_s}$ and $\{\va_{i}^{u}\}_{i=1}^{C_u}$, respectively.
$\va_{i}$ represents the semantic vector for the $i$-th class. In the setting of GZSL, we only have access to the seen samples $\mathcal{S}$ and semantic vectors $\{\va_{i}^{s}\}_{i=1}^{C_s}$ during training. Hence, for brevity, in the demonstration of the training process, we will omit the superscript $s$ for all the seen samples, \textit{i.e.}, $\vx=\vx^s$, $\va = \va^s$.

\subsection{Mining Boundary Samples with Entropy Maximization}
To mitigate the variance collapse issue, we propose to explicitly learn the intra-class relationship by mining boundary samples with entropy maximization.
We start by learning a contrastive network $g(\cdot)$ from seen classes. A visual sample $\vx$ is fused with each semantic vector $\va$ of seen classes as the visual-semantic pairs, and the corresponding one-hot label is formulated as:
\begin{equation}
\begin{gathered}
  \text{OneHot}(\vx, \va_i) = 
\begin{cases}
0, \quad y(\vx) \neq y(\va_i)
\\
1, \quad y(\vx) = y(\va_i)
\end{cases}
 \va_i \in \mathcal{A}^s,
\end{gathered}
\end{equation}
where $y(\cdot)$ denotes the class label of the input. When the labels of visual and the semantic pairs are matched, we assign $1$ to the index of this pair. The contrastive network takes input from the visual-semantic pairs and produces a matching probability between 0 and 1 with a Sigmoid activation function. The classification loss function for the contrastive network can then be formulated as:
\begin{flalign}
\label{g}
\mathcal{L}_{con} &= \sum_{i=1}^{C_s} \lVert g(\vx,\va_{i}) - \text{OneHot}(\vx,\va_{i})\rVert^2,
\end{flalign}
where applies a mean squared error between the matching probability of each pair and the one-hot label.

The aim of mining the boundary samples is to perturb the underlying data distribution so that the predictive uncertainty of the contrastive model is enlarged. Given a visual sample $\vx$ paired with each semantic vector of seen classes, the learned contrastive network $g(\cdot)$ will produce a probability vector over seen classes. This motivates us to involve prediction entropy $H(\vx)$ of all possible classes:
 \begin{flalign} 
\label{h}
H(\vx) = \sum_{i=1}^{C_s} g(\vx, \va_i)\log g(\vx, \va_i).
\end{flalign}

To mine the boundary samples, we fixed the weights in the contrastive network and take the training samples as the learning parameters instead.
By maximizing the prediction entropy while still minimizing the classification loss function, the training samples move towards the decision boundaries against negative classes. The update step can be defined as:
 \begin{flalign}
 \label{update}
\vx^{k+1} \leftarrow \vx^{k} + \eta \bigtriangledown_{\vx} (\mathcal{L}_{con} - \lambda_{1} H(\vx)),
\end{flalign}
where $k$ represents the $k$-th step in total $K$ steps of the training. The perturbed visual samples in the final step are denoted as $\vx^+$ and appended to the original dataset. 

To further diversify the visual samples, we propose the visual perturbation strategy by adding a small amount of noise to these visual features. We begin with sampling a perturbation vector $\ve \in \mathbb{R}^{d_v}$ from a Gaussian distribution with the same size $d_v$ as the visual features,
$\ve \sim \mathcal{N}(\bm{0},\bm{I})$. A diversified sample can be then yielded by:
\begin{equation}
\begin{gathered}
  \vx^{+} = \vx^{+} + \lambda_2 \ve,
\end{gathered}
\end{equation}
where $\lambda_2$ is the coefficient that controls the degree of perturbation.

While the visual perturbation reveals more diversity, we may unexpectedly incorporate some noisy samples resulting in the distribution shifts. We argue that the prototype of each class should be invariant when introducing diversified samples. Thus, in order to avoid distribution shift, we aim to fix the prototype for each class when perturbing the real samples. The class prototype for $c$-th class is defined as the mean vector of the samples in the same class from the empirical distribution $p_{\bar{X}}$:
\begin{equation}
\begin{gathered}
  \mathbb{E}_{\vx_c \sim p_{\bar{X}}^{c}}[\bar{\vx}_c] = \frac{1}{N_c} \sum_{i=1}^{N_c}\vx_c^i,
\end{gathered}
\end{equation}
where $N_c$ is the sample number of the $c$-th class in the training set.

When generating a visual sample conditioned on $c$-th class with the corresponding semantic vector $\va_c$, the expected mean sample should be close to the class prototype given the prior $\vz$ as the mean vector from the distribution, \textit{i.e.,} all zeros $\bm{0}$:
\begin{equation}
\begin{gathered}
  \mathcal{L}_{proto} = \frac{1}{C_{s}} \sum^{C_{s}}_{c=1} \lVert f^{-1}(\bm{0},\va_{c}) - \mathbb{E}_{\vx_c \sim p_{\bar{X}}^{c}}[\bar{\vx}_c] \rVert^{2},
\label{proto}
\end{gathered}
\end{equation}
where $C_{s}$ is the number of seen classes. 

\subsection{Relative Positioning}
\label{gs}
To preserve the geometric information between different attribute categories in the shared subspace, we introduce the relative positioning technique.
Specifically, instead of representing the class-level semantic information using the raw attributes, we determine three semantic anchors to correspondingly obtain the relative position of each class in the attribute space. Such a relative position is defined as a global semantic vector, which could better aligns the semantic space with the visual one. 


The process of determining the three semantic anchors is described as follows. We begin with constructing a semantic graph with the class-level semantic vectors. The edges $\mathcal{E}$ are defined as the cosine similarities between all semantic vectors:
\begin{equation}
\begin{gathered}
  \mathcal{E}_{ij} = \frac{\va_i\cdot\va_j}{\lVert\va_i \rVert_2 \cdot \lVert\va_j \rVert_2},
\end{gathered}
\end{equation}
where $\mathcal{E}_{ij}$ refers to the similarity between $i$-th class and $j$-th class. Then, for each class, we calculate the sum similarities to all other classes:
\begin{equation}
\begin{gathered}
  \vd_{i} = \sum_{j=1}^{C_s}\mathcal{E}_{ij},
\end{gathered}
\end{equation}
where $C_s$ denotes the total number of seen classes. We define the three semantic anchors $\va_{max}$,  $\va_{min}$, $\va_{med}$ with the highest, lowest, and median sum similarities to other semantic vectors. The global semantic vectors are then acquired by computing the responses from these three semantic anchors.

The dimensionality of visual features is usually much higher than that of semantic vectors, \textit{e.g.,} 2,048 vs. 85 in the AWA dataset. Thus, the generation process can be potentially dominated by visual features. In Section \ref{hpa}, we discuss the impact of the semantic vectors' dimensionality. To avoid this issue, we apply three functions $h_{max}(\cdot),h_{min}(\cdot),h_{med}(\cdot)$ to map the semantic responses to a higher dimension through the implementation with an FC layer and a ReLU activation function. The global semantic vector for each class can then be formulated as:
\begin{equation}
\begin{gathered}
  \va^{g}_{i} =  h_{max}(\va_{i} - \va_{max}) + h_{min}(\va_{i} - \va_{min}) \\ + h_{med}(\va_{i} - \va_{med}).
\end{gathered}
\end{equation}
The global semantic vectors revised by relative positioning are fed into the conditional generation flow as the conditional information.

\begin{figure}[t]
\centering
\includegraphics[width=85mm]{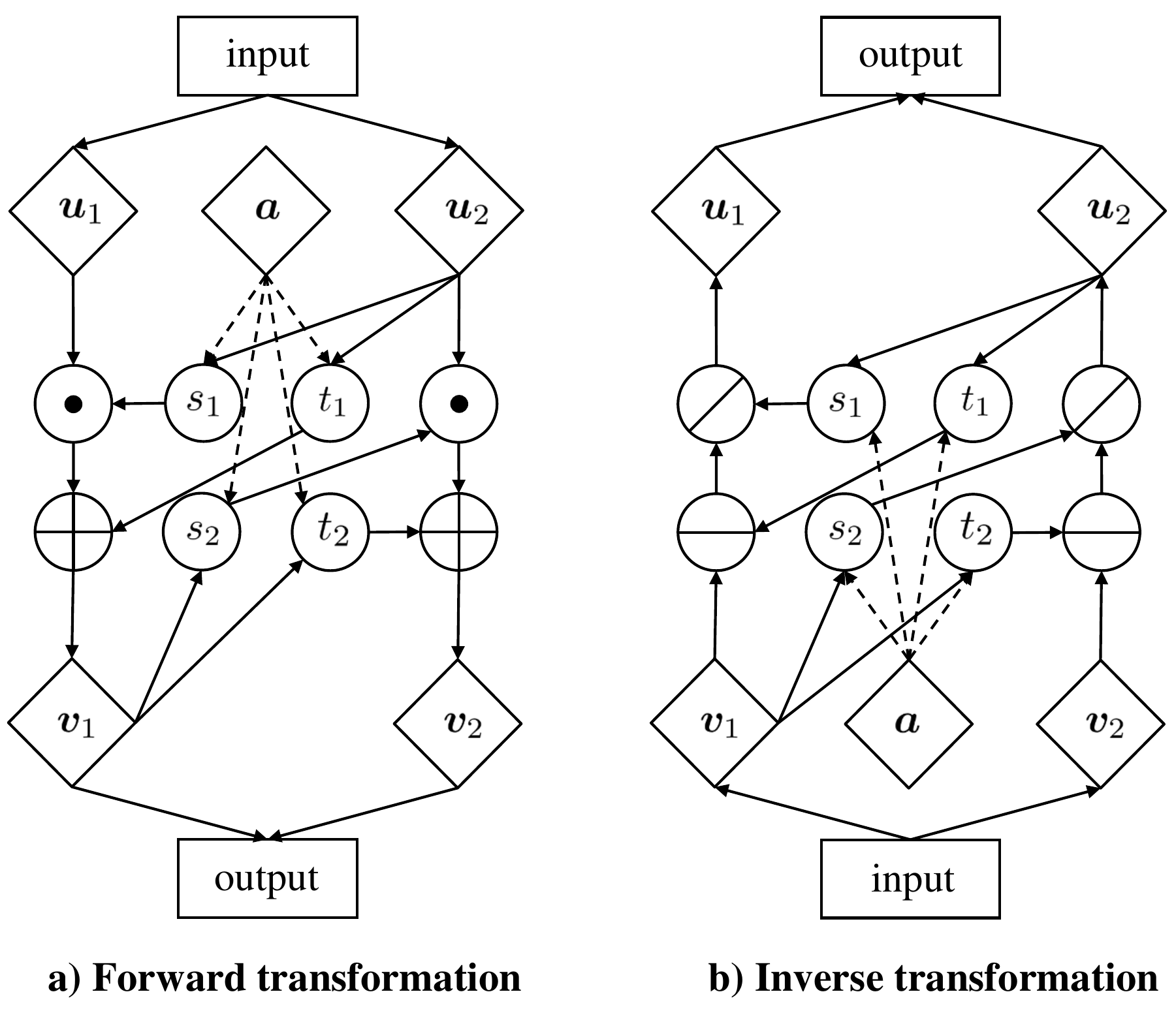}
\caption{The flowcharts of the conditional affine coupling layers. (a) The transformation flowchart when training the model. (b) The inverse direction for generating new samples.}
\label{coupling}
\end{figure}

\subsection{Conditional Generative Flow}
To model the distribution of unseen visual features, we resort to normalizing flow with conditional information, \textit{i.e.,} conditional generative flow \cite{ardizzone2019guided}, a stream of simple yet powerful generative models. 

A conditional generative flow learns inference and generation within the same network. Let $\vz \in \mathcal{Z}$ be a random variable from a particular prior distribution $p_{Z}$, \textit{e.g.,} Gaussian, which has the same dimension as the visual features. We denote the inference process as the forward transformation $f(\cdot): \mathcal{X} \times \mathcal{A} \rightarrow \mathcal{Z}$, whose inverse transformation $f^{-1}(\cdot) : \mathcal{Z} \times \mathcal{A} \rightarrow \mathcal{X}$ is the generation process. The transformation is composed of $L$ bijective transformations $f(\cdot) = f_1(\cdot) \circ f_2(\cdot) \circ... \circ f_L(\cdot)$, the forward and inverse computations are the composition of the $L$ bijective transformations. 
Then, we can formalize the \textit{conditional generative flow} as:
\begin{equation}
\begin{gathered}
  \vz = f(\vx ; \va, \theta), \quad\vz \sim  \mathcal{N}(\bm{0}, \bm{I}),
\end{gathered}
\end{equation}
where the transformation $f(\cdot)$ parameterized with $\theta$ learns to transform a sample $\vx$ in the target distribution to the prior data distribution $p_Z$ conditioned on the corresponding class-level semantic vector $\va$. The forward transformation  has its inverse transformation $f^{-1}(\cdot)$, which flows from the prior distribution $p_Z$ towards the target data distribution $p_X$:
\begin{equation}
\begin{gathered}
  \vx = f^{-1}(\vz; \va, \theta).
\end{gathered}
\end{equation}
With the \textit{change of variable formula}, the bijective function $f(\cdot)$ can be trained through maximum log-likelihood:
\begin{equation}
\begin{gathered}
  \log p_{X}(\vx;\va, \theta) =\log p_{Z}(f(\vx;\va, \theta)) + \log\Big\lvert \text{det}(\frac{\partial f}{\partial \vx} ) \Big\rvert,
\end{gathered}
\end{equation}
where the latter half term denotes the logarithm of the determinant of the Jacobian matrix.

According to Bayes' theorem, the posterior distribution $p_{X}(\theta ; \vx,\va)$ for the parameter $\theta$ is proportional to $p_{X}(\vx;\va,\theta) p_{\theta}(\theta)$. The objective function can then be formulated as:
\begin{equation}
\begin{gathered}
  \mathcal{L}_{flow} = \mathbb{E}[-\log p_{X}(\vx;\va,\theta)] - \log p_{\theta}(\theta),
  \label{flow}
\end{gathered}
\end{equation}

\begin{figure*}[t]
\centering
\includegraphics[width=0.9\textwidth]{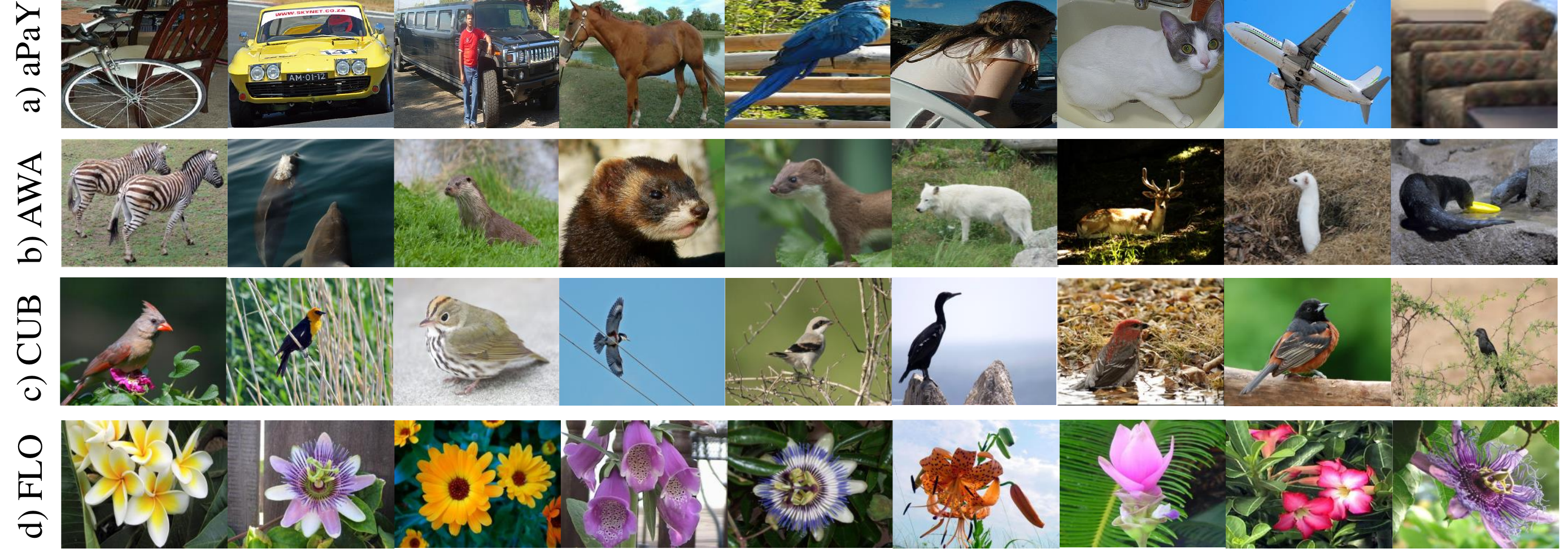}
\caption{Exemplars of four datasets, including two coarse-grained datasets (\textit{i.e.,} aPaY and AWA), and two fine-grained datasets (\textit{i.e.,} CUB and FLO).}
\label{dataset}
\vspace{-15pt}
\end{figure*}

The composed bijective transformations are $L$ conditional affine coupling layers. As shown in Figure \ref{coupling}, in the forward transformation, each layer splits the input vector $\vu$ into two factors $[\vu_1, \vu_2]$. Note that in the first coupling layer, the input is $\vx$. Within each coupling layer, the internal functions $\vs_1(\cdot)$, $\vs_2(\cdot)$, and $\vt_1(\cdot)$, $\vt_2(\cdot)$ are formulated as:
\begin{equation}
\begin{gathered}
  \vv_1 = \vu_1 \odot \text{exp}\big(s_1([\vu_2,\va])\big) + t_1([\vu_2,\va]), \\
  \vv_2 = \vu_2 \odot \text{exp}\big(s_2([\vv_1,\va])\big) + t_2([\vv_1,\va]),
\end{gathered}
\end{equation}
where $\odot$ is the element-wise multiplication and the outputs $\vv_1$ and $\vv_2$ are fed into the next affine transformation.  In the last coupling layer, the output will be $\vz$. In the inverse direction, the conditional affine coupling module takes $\vv_1$ and $\vv_2$ as inputs:
\begin{equation}
    \begin{split}
    &\vu_2 = \big(\vv_2 - t_2([\vv_1,\va])\big) \oslash \text{exp}\big(s_2([\vv_1,\va])\big), \\
    &\vu_1 = \big(\vv_1 - t_1([\vu_2,\va])\big) \oslash \text{exp}\big(s_1([\vu_2,\va])\big),
    \end{split}
\end{equation}
where $\oslash$ is the element-wise division. \noindent As proposed in Real NVP \cite{dinh2016density}, when combining coupling layers, the Jacobian determinant remains tractable and its logarithm is the sum of $s_1(\cdot)$ and $s_2(\cdot)$ over visual feature dimensions.

\subsection{Training and Zero-shot Inference}
We have demonstrated the strategies for improving visual and semantic spaces, and also the conditional generative flow. In this subsection, we will illustrate the training strategy and zero-shot inference stage.

It is worth mentioning that GSMFlow framework involves two training stages. 
First of all, in the boundary sample mining stage, a contrastive network $g{(\cdot)}$ is trained with the loss function defined in Eq. \ref{g}. Then, we free the contrastive network and propagate the original visual samples according to Eq. \ref{update}. The original samples after prorogation for a particular number of steps are the mined boundary samples $\vx^{+}$. 
With the derived boundary samples $\vx^{+}$ and the global semantic vectors $\va^{g}$, the objective functions $\mathcal{L}_{flow}$ and $\mathcal{L}_{proto}$ in Equation \ref{flow} and Equation \ref{proto} should be rewritten as:
\begin{equation}
\begin{gathered}
\mathcal{L}_{flow} = \mathbb{E}[-\log p_{X}(\vx^{+};\va^{g},\theta)] - \log(p_{\theta}(\theta)), \\
\mathcal{L}_{proto} = \frac{1}{C_{s}} \sum^{C_{s}}_{c=1} \lVert f^{-1}(\bm{0},\va^{g}_{c}) - \mathbb{E}_{\vx_c \sim p_{\bar{X}}^{c}}[\bar{\vx}_c] \rVert^{2}.
 \end{gathered}
\end{equation}

Then, the overall objective function of the proposed GSMFlow is formulated as:
\begin{equation}
\begin{gathered}
  \underset{\theta}{min}( \mathcal{L}_{flow} + \lambda_3 \mathcal{L}_{proto}),
\end{gathered}
\end{equation}
where $\lambda_3$ is the coefficient of the prototype loss.

After the conditional generative flow is trained on the seen classes with the boundary samples and the global semantic vectors, it is leveraged to generate visual features of unseen classes:
\begin{equation}
\begin{gathered}
\bar{\vx}^{u} = f^{-1}(\vz;\va^g, \theta).
\end{gathered}
\end{equation}
A softmax classifier is then trained on the real visual features of seen classes and the synthesized visual features of unseen classes. For the coming test samples from either seen or unseen classes, the softmax classifier aims to predict the corresponding class label accurately.

\begin {table}
\caption {Dataset statistics for aPaY, AWA, CUB and FLO. }
\begin{center}
\scalebox{1.1}{
\begin{tabular}{lccccc}
\toprule
 \textbf{Dataset}             & aPaY          & AWA       & CUB   & FLO    \\
\midrule
\#  Attributes     & 64           & 85        & 1024    & 1024   \\
\#  Seen Classes   & 20           & 40        & 150    & 82     \\
\#  Unseen Classes & 12           & 10        & 50     & 20     \\
\#  Total Images   & 18,627       & 30,475    & 11,788 & 8,189  \\ 
\bottomrule 
\specialrule{.1em}{-.15em}{.00em}
\end{tabular}}
\end{center}
\label{dataset}
\end{table}










\begin {table*}[t]
\caption {Performance comparison in accuracy (\%) of the state-of-the-art ZSL and GZSL on four datasets. For ZSL, performance results are reported with the average top-1 classification accuracy (T1). For GZSL, results are reported in terms of top-1 accuracy of unseen (U) and seen (S) classes, together with their harmonic mean (H). The best results of the harmonic mean are highlighted in bold. $\dag$ and $\ddag$ represent embedding-based and generative methods, respectively.$*$ indicates the performance results of method in our conference version. For this extension, we report the average performance results and the standard deviations from multiple experiments with different random seeds.
}
\begin{center}
\scalebox{0.95}{
\begin{tabular}[t]{c|c|cccc|cccc|cccc|cccc}

 \hline 
\noalign{\smallskip}

 \multirow{2}{*}{} &  \multirow{2}{*}{}   & \multicolumn{4}{c|}{aPaY}   & \multicolumn{4}{c|}{AWA}  &  \multicolumn{4}{c|}{CUB} & \multicolumn{4}{c}{FLO}  \\

  & \multirow{-2}{*}{Methods}  & \multicolumn{1}{c|}{\textit{T1}} &\textit{U} & \textit{S} & \textit{H}  & \multicolumn{1}{c|}{\textit{T1}} & \textit{U} & \textit{S} & \textit{H}  & \multicolumn{1}{c|}{\textit{T1}} & \textit{U} & \textit{S} & \textit{H} & \multicolumn{1}{c|}{\textit{T1}} & \textit{U} & \textit{S} & \textit{H}\\

  \noalign{\smallskip}
 \hline 
\noalign{\smallskip}
\multirow{6}{*}{\dag}   
                      &   LATEM   \cite{xian2016latent}  
&\multicolumn{1}{c|}{35.2}    & 0.1           & 73.0              & 0.2        
&\multicolumn{1}{c|}{55.8}    & 13.3          & 77.3              & 20.0
&\multicolumn{1}{c|}{49.3}    & 15.2          & 57.3              & 24.0    
&\multicolumn{1}{c|}{60.8}    & 6.6           & 47.6              & 11.5  
    \\
 
                      &   ALE    \cite{akata2015label}   
&\multicolumn{1}{c|}{39.7}    & 4.6           & 73.7              & 8.7      
&\multicolumn{1}{c|}{62.5}    & 14.0          & 81.8              & 23.9    
&\multicolumn{1}{c|}{54.9}    & 23.7          & 62.8              & 34.4    
&\multicolumn{1}{c|}{48.5}    & 13.3          & 61.6              & 21.9  
    
    \\
                   &   SJE       \cite{akata2015evaluation}
&\multicolumn{1}{c|}{31.7}    & 1.3          & 71.4              & 2.6       
&\multicolumn{1}{c|}{61.9}    & 8.0          & 73.9              & 14.4  
&\multicolumn{1}{c|}{54.0}    & 23.5         & 59.2              & 33.6      
&\multicolumn{1}{c|}{53.4}    & 13.9         & 47.6              & 21.5    
    \\

                    & SAE     \cite{kodirov2017semantic}  
&\multicolumn{1}{c|}{8.3}    & 0.4           & 80.9              & 0.9        
&\multicolumn{1}{c|}{54.1}    & 1.1           & 82.2              & 2.2 
&\multicolumn{1}{c|}{33.3}    & 7.8           & 54.0              & 13.6
&\multicolumn{1}{c|}{-}    & -             & -                 & -    
    \\

                    & TCN       \cite{jiang2019transferable}
&\multicolumn{1}{c|}{38.9}    & 24.1          & 64.0              & 35.1       
&\multicolumn{1}{c|}{71.2}    & 61.2          & 65.8              & 63.4
&\multicolumn{1}{c|}{59.5}    & 52.6          & 52.0              & 52.3      
&\multicolumn{1}{c|}{-}    & -             & -                 & -    
    \\
                    & DVBE      \cite{min2020domain}
&\multicolumn{1}{c|}{-}    & 32.6          & 58.3              & 41.8          
&\multicolumn{1}{c|}{-}    & 63.6          & 70.8              & 67.0     
&\multicolumn{1}{c|}{-}    & 53.2          & 60.2              & 56.5         
&\multicolumn{1}{c|}{-}    & -             & -                 & -   
    \\

 \noalign{\smallskip}
 \hline 
\noalign{\smallskip}
\multirow{10}{*}{\ddag} & GAZSL \cite{zhu2018generative}      
&\multicolumn{1}{c|}{41.1}    & 14.2          & 78.6              & 24.0      
&\multicolumn{1}{c|}{70.2}    & 35.4	        & 86.9              & 50.3   
&\multicolumn{1}{c|}{55.8}    & 31.7	        & 61.3	            & 41.8       
&\multicolumn{1}{c|}{60.5}    & 28.1          & 77.4              & 41.2 
    \\

                        & f-CLSWGAN \cite{xian2018feature}     
&\multicolumn{1}{c|}{40.5}   & 32.9           & 61.7              & 42.9    
&\multicolumn{1}{c|}{65.3}   & 56.1           & 65.5              & 60.4     
&\multicolumn{1}{c|}{57.3}   & 43.7           & 57.7              & 49.7	       
&\multicolumn{1}{c|}{69.6}   & 59.0           & 73.8              & 65.6 
   \\
                     &  CANZSL   \cite{chen2020canzsl}         
&\multicolumn{1}{c|}{-}    & -                 & -                 & -  
&\multicolumn{1}{c|}{68.9}    & 49.7              & 70.2              & 58.2  
&\multicolumn{1}{c|}{60.6}    & 47.9              & 58.1              & 52.5 
&\multicolumn{1}{c|}{69.7}    & 58.2              & 77.6              & 66.5 
    \\
 
                    & CADA-VAE  \cite{schonfeld2019generalized}
&\multicolumn{1}{c|}{42.3}    & 31.7              & 55.1              & 40.3     
&\multicolumn{1}{c|}{64.0}    & 55.8              & 75.0              & 63.9     
&\multicolumn{1}{c|}{60.4}    & 51.6              & 53.5              & 52.4     
&\multicolumn{1}{c|}{65.2}    & 51.6              & 75.6              & 61.3 
    \\        
                    & f-VAEGAN-D2  \cite{xian2019f} 
&\multicolumn{1}{c|}{-}    & -                 & -                 & -         
&\multicolumn{1}{c|}{71.1}    & 57.6              & 70.6              & 63.5     
&\multicolumn{1}{c|}{61.0}    & 48.4              & 60.1              & 53.6     
&\multicolumn{1}{c|}{67.7}    & 56.8              & 74.9              & 64.6 
    \\

                    & TF-VAEGAN  \cite{narayan2020latent}
&\multicolumn{1}{c|}{-}    & -                 & -                 & -     
&\multicolumn{1}{c|}{72.2}    & 59.8              & 75.1              &  66.6  
&\multicolumn{1}{c|}{64.9}    & 52.8              & 64.7              & 58.1       
&\multicolumn{1}{c|}{70.8}    & 62.5              & 84.1              & 71.7  
    \\

                    & E-PGN  \cite{yu2020episode}
&\multicolumn{1}{c|}{-}    & -                 & -                 & -          
&\multicolumn{1}{c|}{73.4}    & 52.6              & 83.5              & 64.6    
&\multicolumn{1}{c|}{72.4}    & 52.0              & 61.1              & 56.2 
&\multicolumn{1}{c|}{85.7}    & 71.5              & 82.2              & 76.5  
     
    \\ 
        & IZF \cite{shen2020invertible}
&\multicolumn{1}{c|}{44.9}    & 42.3              & 60.5              & 49.8
&\multicolumn{1}{c|}{\textbf{74.5}}    & 60.6              & 77.5              & 68.0    
&\multicolumn{1}{c|}{67.1}    & 52.7              & 68.0              & 59.4    
&\multicolumn{1}{c|}{-}    & -                 & -                 & -  
     
    \\ 
          & OOD-GZSL \cite{chen2020boundary}
&\multicolumn{1}{c|}{-}    & -                 & -                 & -          
&\multicolumn{1}{c|}{-}    & 55.9              & 94.9              & 70.3    
&\multicolumn{1}{c|}{-}    & 53.8              & 94.6             &\textbf{68.6}
&\multicolumn{1}{c|}{-}    & 61.9              & 91.7              & 73.9  
     
    \\  
                & CE-GZSL \cite{han2021contrastive}
&\multicolumn{1}{c|}{-}    & -                 & -                 & -          
&\multicolumn{1}{c|}{70.4}   & 63.1              & 78.6              & 70.0    
&\multicolumn{1}{c|}{77.5}    & 63.9              & 66.8             &{65.3}
&\multicolumn{1}{c|}{70.6}    & 69.0              & 78.7              & 73.5  
     
    \\  
    
     & FREE \cite{chen2021free}
&\multicolumn{1}{c|}{-}   & -                 & -                 & -          
&\multicolumn{1}{c|}{-}    & 60.4              & 75.4              & 67.1    
&\multicolumn{1}{c|}{-}    & 55.7              & 59.9              & 57.7
&\multicolumn{1}{c|}{-}    & 67.4              & 84.5              & 75.0  
     \\
          & SDGZSL \cite{chen2021iccv}
&\multicolumn{1}{c|}{45.4}    & 38.0              & 57.4              & 45.7          
&\multicolumn{1}{c|}{72.1}    & 64.6              & 73.6              & 68.8    
&\multicolumn{1}{c|}{75.5}    & 59.9              & 66.4         &63.0
&\multicolumn{1}{c|}{{86.9}}    & 83.3              & 90.2             & 86.6  
     
    \\  
    
\noalign{\smallskip}
 \hline 
\noalign{\smallskip}
 &   GSMFlow$*$
&\multicolumn{1}{c|}{{49.2}} & 42.0 & 62.3 & {50.2}  
&\multicolumn{1}{c|}{72.7} & 64.5 & 82.1 & {72.3}
&\multicolumn{1}{c|}{{76.4}} & 61.4 & 67.4 & 64.3
&\multicolumn{1}{c|}{{86.9}} & 86.6 & 87.8 & {87.2} 
\\
    
\noalign{\smallskip}
 \hline 
\noalign{\smallskip}
 &  \multirow{2}{*}{\textbf{GSMFlow}}  
&\multicolumn{1}{c|}{\textbf{50.7}} & 40.9 & 68.9 & \textbf{51.3}  
&\multicolumn{1}{c|}{73.2} & 66.5 & 80.3 & \textbf{72.8}
&\multicolumn{1}{c|}{\textbf{78.3 }} & 63.2 & 69.3 & 66.1
&\multicolumn{1}{c|}{\textbf{90.6}} & 86.9 & 89.8 & \textbf{88.3} 
\\
[-4pt]
&   
&\multicolumn{1}{c|}{\tiny{{$\pm 1.2$}}}
& \multicolumn{1}{c}{\tiny{{$\pm 0.6$}}}
& \multicolumn{1}{c}{\tiny{{$\pm 1.1$}}}
& \multicolumn{1}{c|}{\tiny{{$\pm 0.7$}}}

& \multicolumn{1}{c|}{\tiny{{$\pm 0.3$}}}
& \multicolumn{1}{c}{\tiny{{$\pm 0.7$}}}
& \multicolumn{1}{c}{\tiny{{$\pm 0.7$}}}
& \multicolumn{1}{c|}{\tiny{{$\pm 0.5$}}}

& \multicolumn{1}{c|}{\tiny{{$\pm 0.2$}}}
& \multicolumn{1}{c}{\tiny{{$\pm 0.7$}}}
& \multicolumn{1}{c}{\tiny{ {$\pm 0.7$}}}
& \multicolumn{1}{c|}{\tiny{ {$\pm 0.2$}}}

&\multicolumn{1}{c|}{\tiny{ {$\pm 0.5$}}}
&\multicolumn{1}{c}{\tiny{ {$\pm 0.3$}}}
&\multicolumn{1}{c}{\tiny{ {$\pm 0.4$}}}
&\multicolumn{1}{c}{\tiny{ {$\pm 0.2$}}}

\\
\noalign{\smallskip}
 \hline 
\end{tabular}}
\end{center}
\label{gzslperoformance}
\end {table*}

\section{Experiments}
In this section, we evaluate our approach GSMFlow in both generalized zero-shot learning and conventional zero-shot learning tasks. We first introduce the datasets and experimental settings and then compare GSMFlow with the state-of-the-art methods. Finally, we study the effectiveness of the proposed model with a series of ablation study and hyper-parameter sensitivity analysis. 

\subsection{Datasets}
We conduct experiments on four widely used benchmark datasets of image classification. They are two fine-grained datasets, \textit{i.e.,} Caltech-UCSD Birds-200-2011 (\textbf{CUB}) \cite{wah2011caltech} and Oxford  Flowers (\textbf{FLO}) \cite{nilsback2008automated}, two coarse-grained datasets, \textit{i.e.,} Attribute Pascal and Yahoo (\textbf{aPaY}) \cite{farhadi2009describing} and Animals with Attributes 2 (\textbf{AWA}) \cite{lampert2013attribute}.
CUB consists of 11,788 images from 200 fine-grained bird species, in which 150 selected as seen classes and 50 as unseen classes. For FLO, it contains 8,189 images from 102 flower categories, 82 of which are chosen as seen classes. The class-level semantic vectors of CUB and FLO are extracted from the fine-grained visual descriptions (10 sentences per image), yielding 1,024-dimensional character-based CNN-RNN features \cite{reed2016learning} for each class. aPaY dataset contains 18,627 images from 42 classes. There are 30 seen classes and 12 unseen classes respectively. Each class is annotated with 64 attributes. AWA2 is a considerably larger dataset with 30,475 images from 50 classes and they are annotated with 85 attributes. 

Following the compared methods, we used ResNet101 pretrained on ImageNet1k  without finetuning to extract the 2048d visual features from the entire images.
For the split, 40 of the total classes are selected as seen classes and ten as unseen classes. For each dataset, We follow the proposed split setting in \cite{xian2018zero} and \cite{nilsback2008automated}. 

\subsection{Implementation Details}
Our framework is implemented with the open-source machine learning library PyTorch. The conditional generative flow consists of a series of affine coupling layers. Each affine coupling layer is implemented with two fully connected (FC) layers and the first FC layer is followed by a LeakyReLU activation function. The hidden dimension of the FC layer is set as 2,048. The coefficients $\lambda_1$ of perturbation degree and the coefficient $\lambda_2$ of the prototype loss are set within \{0.02, 0.05, 0.15, 0.3, 0.5\} and \{1, 3, 10, 20, 30\}. The dimension of global semantic vectors varies in \{128, 256, 512, 1,024, 2,048\} and the number of the affine coupling layers varies in \{1, 3, 5, 10, 20\}.  The corresponding results are given in Section \ref{hpa}. The function $h(\cdot)$ is implemented with an FC layer and a ReLU activation function. 
The contrastive network $g(\cdot)$ for mining the boundary samples is implemented by two linear layers, following with a ReLU activation function and a Sigmoid activation function, respectively. 
We use Adam optimizer with $\beta_{1}=0.9$, $\beta_{2}=0.999$ and set the batch size to 256 and set the learning rate to 3e-4. All the experiments are performed on a Lenovo workstation with two NVIDIA GeForce GTX 2080 Ti GPUs.

\begin{figure*}[t]
\centering
\includegraphics[width=0.95\textwidth]{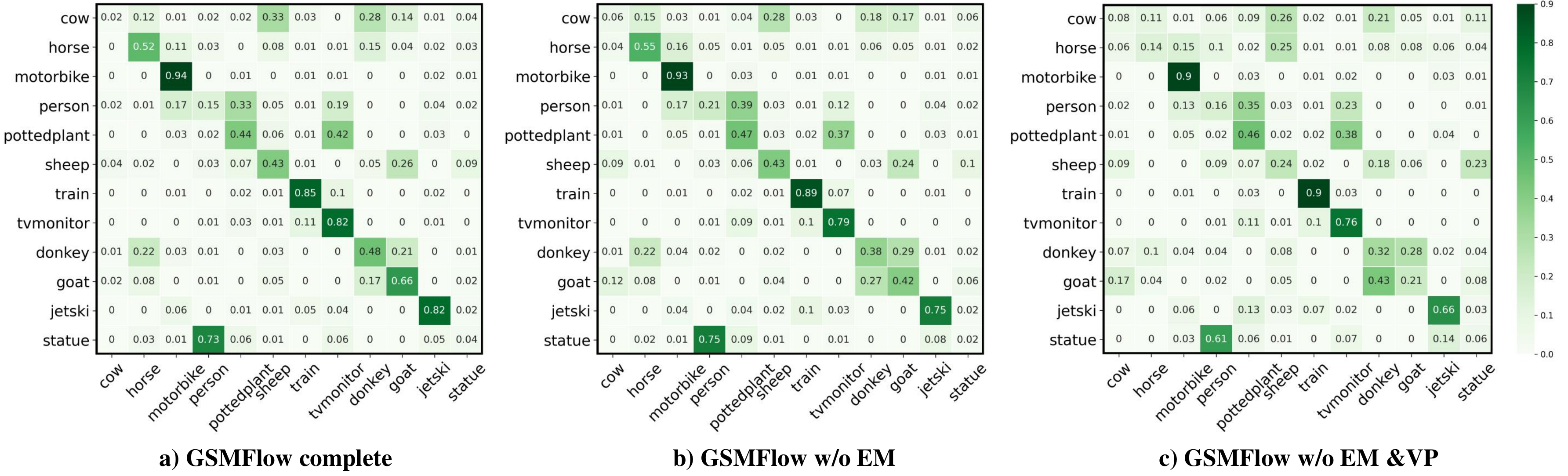}
\caption{Class-wise performance comparison between with and without visual perturbations. The vertical labels are the groundtruth and the horizontal labels are the predictions.}
\label{cm}
\end{figure*}

\subsection{Evaluation Metrics}
To avoid the failure of classification accuracy for imbalanced class distributions, we adopt average per-class Top-1 accuracy as the fair evaluation criteria for conventional ZSL and the seen and unseen set performance in GZSL:
\begin{equation}
\begin{aligned}
  Acc_{\mathcal{Y}} = \frac{1}{|\mathcal{Y}|} \sum^{|\mathcal{Y}|}_{y=1} \frac{\# \  of \ correct \ predictions \ in \ y}{\# \ of \ samples \ in \ y},
\end{aligned}
\end{equation}
where $|\mathcal{Y}|$ is the number of testing classes. A correct prediction is defined as the highest probability of all candidate classes.
Following \cite{xian2018zero}, the harmonic mean of the average per-class Top-1 accuracies on seen $Acc_S$ and unseen $Acc_U$ classes are used to evaluate the performance of generalized zero-shot learning. It is computed by:
\begin{equation}
\begin{aligned}
  H = \frac{2*Acc_S*Acc_U}{Acc_S+Acc_U}.
\end{aligned}
\end{equation}

\subsection{Comparisons with State-of-the-art Methods}
\label{GZSL}

Table \ref{gzslperoformance} summarizes the performance comparison between our proposed GSMFlow and the other state-of-the-art methods in the setting of GZSL. We choose the most representative state-of-the-art methods for comparison, the top six methods marked with $\dag$ are embedding-based methods, while the below eleven methods and our proposed method marked with $\ddag$ are generative methods. The unseen, seen and the balanced harmonic mean between the two domains are represented as U, S, and H. It can be seen that our proposed framework consistently outperforms other methods except on CUB dataset. 
For the semantic information of CUB in the compared methods, the recent methods E-PGN, CE-GZSL and SDGZSL also use the 1,024-dimensional semantic information as in this work.
 It is worth noting that  OOD-ZSL achieves the best performance on CUB as it leverages a specific calibration technique to threshold the confidence of the prediction being seen or unseen. The thresholding hyperparameter is sensitive to the datasets, which makes it hard to generalize when more unseen classes are involved. Among the generative methods, IZF \cite{shen2020invertible} also leverages normalizing flows as the base generative model.
However, we show that by mitigating the generation shifts, our proposed GSMFlow achieves significant improvements on these four datasets. In addition to the conference version, the proposed boundary sample mining technique with entropy maximization introduces more diversified visual potentials to enrich the visual space. In result, we achieve 1.1\%, 0.5\%, 1.8\% and 1.1\% higher on the four datasets respectively.


\subsection{Conventional Zero-shot Learning}
\label{ZSL}
Even if GSMFlow is mainly proposed for GZSL, to further validate the effectiveness of our proposed framework, we also conduct experiments in the context of conventional ZSL. Table \ref{gzslperoformance} summarizes the conventional zero-shot learning performance, which is reported under the T1 metric. We achieve better performance than all the compared methods on the aPaY, CUB, and FLO datasets. For the AWA dataset, we can also achieve comparatively good performance. It can be seen that even if IZF achieves the best performance on AWA dataset with the conventional zero-shot learning setting, i.e., only considering the unseen classes during inference. Our proposed GSMFlow shows a significant advantage on the generalized zero-shot learning setting, where both seen and unseen classes are considered. Specifically, GSMFlow surpass IZF by 5.9\%, 2.8\% and 4.8\% on U, S and H respectively. Comparing with the performance results reported in our conference version, the proposed boundary sample mining technique boosts the conventional ZSL performance on the four datasets by 1.5\%, 0.5\%, 1.9\%, and 3.7\% respectively.



\subsection{Class-wise Performance Comparison}
To further investigate the performance boost through boundary samples and visual perturbation, in Figure \ref{cm}, we compare the class-wise accuracy between \textbf{a) GSMFlow complete}, \textbf{b) GSMFlow w/o EM} and \textbf{c) GSMFlow w/o EM\&VP} by illustrating the confusion matrices in the three settings on aPaY dataset. 
Comparing between \textbf{b)} and \textbf{c)}, when perturbing the visual features, the generator is exposed to more diverse visual samples, resulting in better generalization ability to unseen classes. Especially, in this case study, the \textit{horses} and the \textit{sheep} are the two unseen classes in aPaY dataset. We can notice that \textit{horses} can be easily misclassified as \textit{sheep} and can only achieve 14\% accuracy. The performance surges to 55\% when we introduce the visual perturbation strategy. Similar observations hold for other classes. We further compare class-wise performance between with and without mining boundary samples. It can be seen that the performance improvement mostly come from three classes, \textit{i.e.,} \textbf{donkey}, \textbf{goat}, and \textbf{jetski}. By learning from boundary samples, we surpass the model in the conference version by 10\%, 24\% and 7\% respectively.

\begin {table*}[t]
\caption {Effects of different components on four datasets. \textit{U}, \textit{S} and \textit{H} represent unseen, seen and harmonic mean, respectively. The best results of harmonic mean are formatted in bold.}
\begin{center}
\scalebox{1.08}{
\begin{tabular}[t]{c|ccc|ccc|ccc|ccc}

\hline 
\noalign{\smallskip}

       & \multicolumn{3}{c|}{aPaY} & \multicolumn{3}{c}{AWA}  & \multicolumn{3}{c|}{CUB} & \multicolumn{3}{c}{FLO} \\
  \multirow{-2}{*}{Methods}  & \textit{U} & \textit{S} & \textit{H}  & \textit{U} & \textit{S} & \textit{H}   & \textit{U} & \textit{S} & \textit{H}  & \textit{U} & \textit{S} & \textit{H}  \\
\noalign{\smallskip}
\hline 
\noalign{\smallskip}

  IZF w/o constraints      
            & 35.2      & 54.2      & 42.7      
            & 38.1      & 78.9      & 51.4   
            & 48.6      & 54.3      & 51.3
            & 59.3      & 78.2      & 67.5
            \\

  GSMFlow w/o constraints   
            & 36.5      & 60.0      & 45.4       
            & 53.3      & 67.6      & 59.6   
            & 52.6      & 61.2      & 56.6
            & 68.6      & 80.1      & 73.9
            \\

  GSMFlow w/o EM\&VP     
            & 38.4      & 62.3      & 47.6        
            & 62.2      & 71.3      & 66.4   
            & 57.9      & 65.2      & 61.3
            & 81.3      & 83.8      & 82.5
            \\

 GSMFlow w/o EM\&RP
            & 39.6      & 61.3      & 48.1         
            & 62.9      & 80.4      & 70.6
            & 56.8      & 66.2      & 61.1
            & 80.9      & 86.2      & 83.4
            \\
 GSMFlow w/o EM           
& 42.0 & 62.3 & {50.2}  
& 64.5 & 82.1 & {72.3}
& 61.4 & 67.4 & 64.3 
&86.6  & 87.8 & 87.2 \\
  {GSMFlow w/o VP}           
&  {39.0} &  {64.8} &  {48.7}  
&  {64.8} &  {81.2} &  {72.1}
&  {62.1} &  {69.2} &  {65.4} 
&  {86.5} &  {88.8} &  {87.6} \\
\noalign{\smallskip}
\hline 
\noalign{\smallskip}

 GSMFlow      
 & 40.9 & 68.9 & \textbf{51.3}  
 & 66.5 & 80.3 & \textbf{72.8}
 & 63.2 & 69.3 & \textbf{66.1}
 & 86.9 & 89.8 & \textbf{88.3} 
\\

\noalign{\smallskip}
 \hline 
\end{tabular}}
\end{center}
\label{ablation}
\end {table*}

\begin{figure*}[t]
\centering
\includegraphics[width=0.928\textwidth]{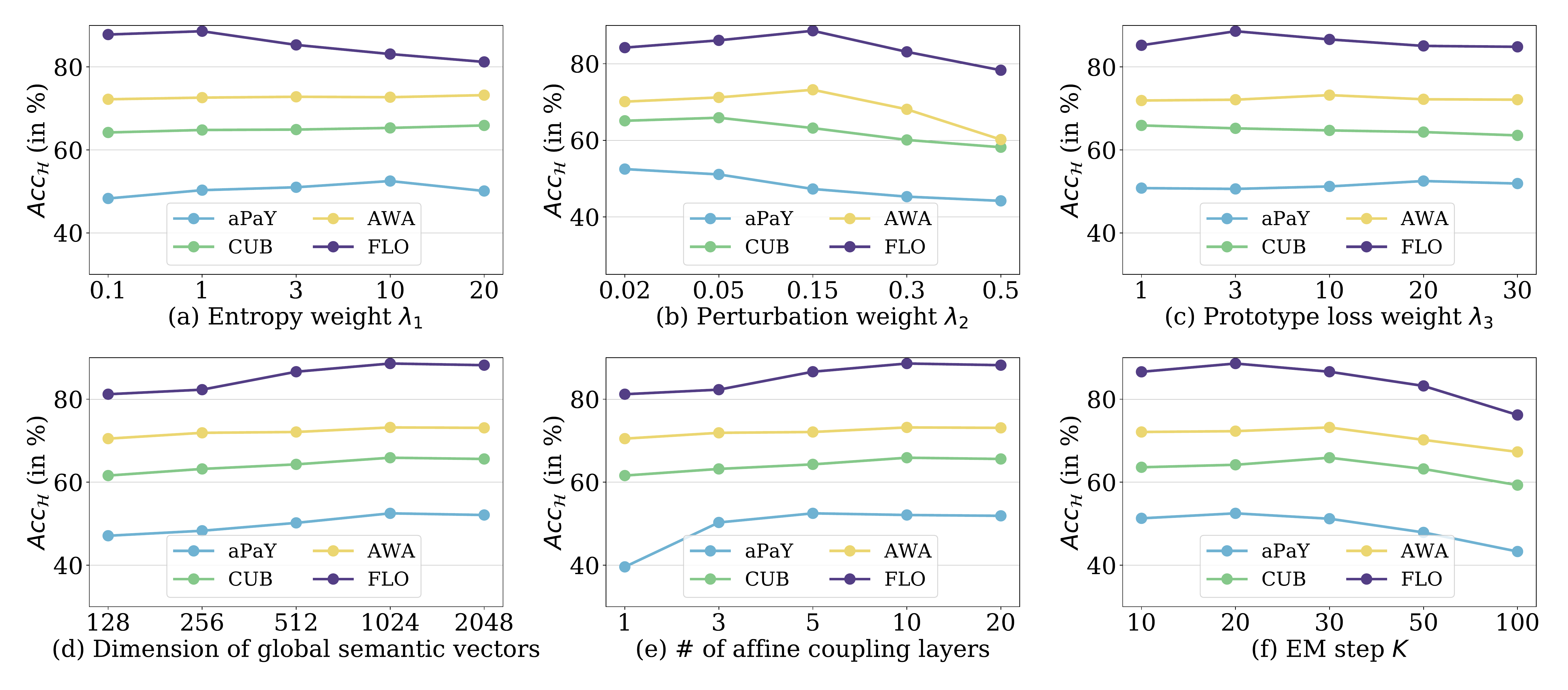}
\caption{Hyper-parameter sensitivity. The horizontal axis indicates the varying hyper-parameters for (a) entropy weight, (b) perturbation weight, (c) prototype loss weight, (d) semantic vector dimensions, (e) number of affine coupling layers and (f) the number of entropy maximization steps. The vertical axis reports the corresponding performance.}
\label{hyper}
\end{figure*}

\subsection{Ablation Study}
To analyze the contribution of each proposed component and the merit from addressing the three problems in generative zero-shot learning, \textit{i.e., semantic inconsistency, variance dacay}, and \textit{structure disorder}, we conduct an ablation study on the proposed GSMFlow. We decompose the complete framework into six variants. These include: \textit{\textbf{IZF w/o constraints}} - the conditional generative flow adopted in compared method \cite{shen2020invertible}; \textit{\textbf{GSMFlow w/o constraints}} - the generative flow in our proposed framework without visual perturbation and global semantic learning; \textit{\textbf{GSMFlow w/o EM\&VP}} - the complete framework without boundary samples and visual perturbation; \textit{\textbf{GSMFlow w/o EM\&RP}} - the original class-level semantic vectors are used and the boundary samples are not introduced;  \textit{\textbf{GSMFlow w/o EM}} - the boundary samples are not introduced (the conference version);  {and \textit{\textbf{GSMFlow w/o VP}} - the boundary samples are introduced but the visual perturbation is removed.}

Comparing between \textit{\textbf{IZF w/o constraints}} and \textit{\textbf{GSMFlow w/o constraints}}, we can see the performance comparison between the two ways of incorporating conditional information in the generative flow. It can be seen that, instead of mixing the conditional information with the prior input in IZF, our explicit conditional strategy that progressively injects the semantic information into the affine coupling layers during training can achieve higher accuracy on GZSL. 
The results on the variants \textit{\textbf{GSMFlow w/o EM\&VP}} shows that by mitigating the structure disorder issue, the relative positioning strategy that captures geometric relationships between semantic vectors can significantly improve the GZSL performance. 
\textit{\textbf{GSMFlow w/o EM\&RP}} indicates that effectiveness of the visual perturbation strategy to prevent the variance collapse. 
\textit{\textbf{GSMFlow w/o EM}} represents the method in the conference version without boundary samples mining. It can be seen that when VP and RP are both applied, significant improvements are observed. 
\textit{\textbf{GSMFlow w/o VP}} shows the effectiveness of the visual perturbation strategy. It can be seen from Table III that when removing the visual perturbation component, the performance results are inferior to the complete framework. This variant setting also demonstrates that the boundary samples and diverse samples are complementary to each other on improving the zero-shot learning performance. 
Lastly, when combining these components together, we achieve the best performance results.

\begin{figure*}[t]
\centering
\includegraphics[width=1\textwidth]{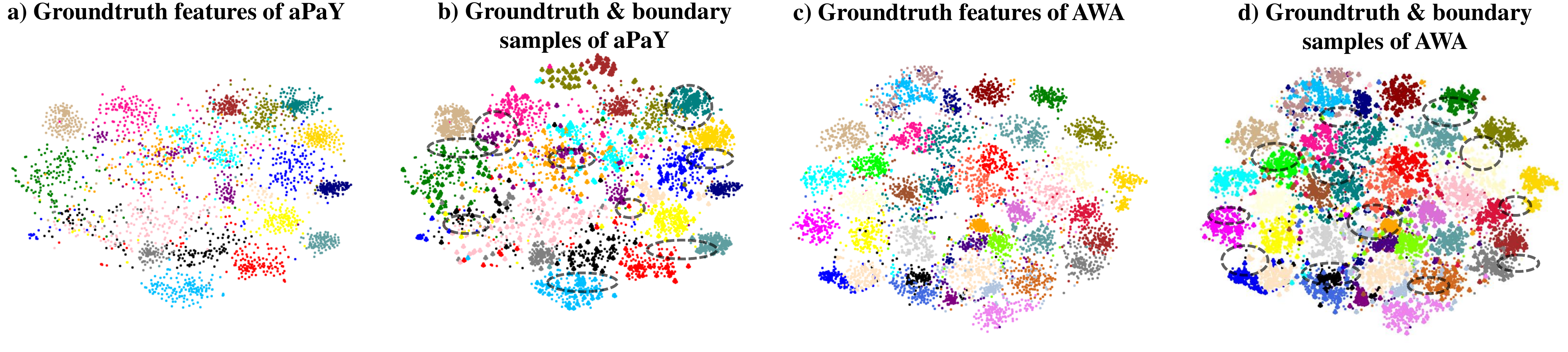}
\caption{t-SNE visualization of groundtruth features and  boundary samples visualization on aPaY and AWA datasets. The dot points represent the groundtruth samples and clubsuit points are the derived boundary samples.}
\label{tsne1}
\end{figure*}

\begin{figure*}[t]
\centering
\includegraphics[width=\textwidth]{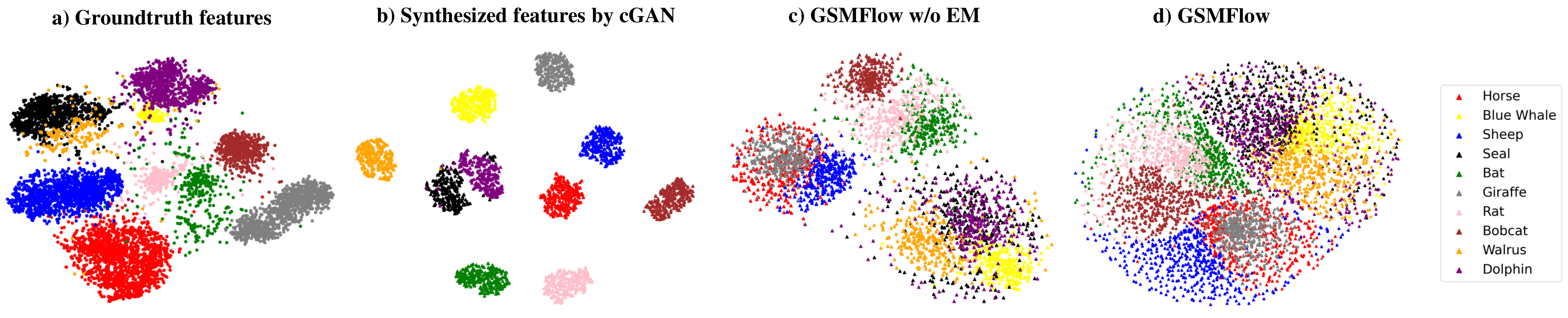}
\caption{Distribution comparison of the unseen classes in the AWA dataset. (a) The real distributions of the visual features extracted from the backbone. (b) The synthesized distributions by cGAN. (c) The synthesized distributions by GSMFlow w/o EM. (d) The synthesized distribution by the complete GSMFlow model.}
\label{tsne2}
\end{figure*}

\subsection{Hyper-parameter Analysis}
\label{hpa}
GSMFlow mainly involves six hyper-parameters in the boundary sample mining and the model training, as shown in Figure \ref{hyper}, including the entropy weight $\lambda_1$, perturbation weight $\lambda_2$, prototype loss weight $\lambda_3$, dimension of the global semantic vectors, number of the conditional affine coupling layers and the entropy maximization 
Varying the entropy weight from 0.1 to 20 for deriving the boundary samples, in Figure \ref{hyper}(a), we find that AWA and CUB both reach the best performance at 20. aPaY and FLO achieve the best performance at 10 and 1, respectively. The entropy weight controls the smoothness of the probability vector. A higher weight for the entropy loss regularizes the probability vector to be more flat. As different flower categories in FLO are visually similar, a small weight for entropy loss is enough to derive the boundary samples.
It turns out that the perturbation weight also varies for different datasets, as shown in Figure \ref{hyper}(b), the accuracies of aPaY, CUB, AWA, and FLO reach the peak at around 0.02, 0.05, 0.15 and 0.15.  {There is an overall trend on all datasets that the performance decreases with larger perturbation weights. A small perturbation weight could explore more diverse samples. However, when the perturbation weight is too large, the features could become indistinguishable between other classes. In general, a small perturbation weight can help to achieve better zero-shot learning performance results.}
The weight of the prototype loss $\lambda_3$ does not have a significant impact on the performance.
In Figure \ref{hyper}(c), we vary $\lambda_3$ between 1 and 30, and observe that at 20, 1, 10, and 3, the best performance results are achieved on aPaY, CUB, AWA and FLO. 
As discussed in Section \ref{gs}, the semantic vectors usually have lower dimensions than the visual features, which makes the visual features dominate the generation process in the affine coupling layers. 
In Figure \ref{hyper}(d), we report the impact from the dimensionality of the semantic vectors. It can be seen that low-dimensional semantic vectors tend to jeopardize the performance. All the best performance results are achieved at 1,024 dimensions. 
We also investigate the impact on the number of conditional affine coupling layers, which directly influences the generative model size. In Figure \ref{hyper}(e), we can see the best performance results are from 5 layers for aPaY and 3 layers for other datasets. Lastly, we experiment on the number of steps for boundary sample mining. As shown in Figure \ref{hyper}(f), updating the training samples for 20 steps is the best for aPaY and FLo and 30 steps for CUB and AWA datasets.

\subsection{t-SNE visualization}
The quantitative results reported for GZSL (Section \ref{GZSL}) and ZSL (Section \ref{ZSL}) demonstrate that the visual samples of the unseen classes generated by GSMFlow are of good quality and effective for classification tasks. 
To further gain an insight into the boundary samples and the quality of the generated samples for validating the motivation illustrated in Figure \ref{intro}, we provide two series of visualization results.

In Figure \ref{tsne1}, we visualize the comparison between the distributions of groundtruth samples without and with boundary samples in (a)(c) and (b)(d) for aPaY and AWA datasets. It can be seen that the derived boundary samples greatly impact the class decision boundaries. In Figure \ref{tsne1} (a) and (c), the inter-class margin is considerably large. After introducing the boundary samples, in Figure \ref{tsne1} (b) and (d) the visual feature space is enriched and inter-class margin is minimized, which represents more extreme visual potentials are explored. The dotted ellipses highlight that the boundary samples that update the decision margin.

We further compare the empirical distribution of the real unseen class data, the synthesized unseen class data by a conditional generative adversarial network (cGAN), our proposed GSMFlow without boundary sampples, and our complete model GSMFlow, as depicted in Figure \ref{tsne2}. 
It can be seen that the distributions of the synthesized unseen data from cGAN suffer from the generation shifts problem, which is undesirable for approximating optimal decision boundaries. Specifically, each class is collapsed to fixed modes without capturing enough variances. 
In contrast, thanks to the visual perturbation strategy, the synthesized samples by GSMFlow are much more diverse.
The class-wise relationship is also reflected. 
For example, comparing to other animal species, \textit{horse} and \textit{giraffe} share some attribute values. Also, all the marine species, \textit{blue whale, walrus, dolphin, and seal} are well separated from other animals. 
Further, in Figure \ref{tsne2} (d) we can see that the margin of decision boundaries are greatly minimized. At the same time, the inter-class decision boundaries are still discriminative. As a result, our generated samples are semantically consistent and more suitable for zero-shot recognition. 


\section{Conclusion}
In this paper, we propose a novel Generation Shifts Mitigating Flow (GSMFlow) framework, which is comprised of multiple conditional affine coupling layers for learning unseen data synthesis. The main motivation of the proposed framework is to address three potential distribution shifts in sample generation, \textit{i.e., semantic inconsistency, variance collapse}, and the \textit{structure disorder}. In detail, we explicitly blend the semantic information into the transformations in each of the coupling layers, reinforcing the correlations between the generated samples and the corresponding attributes. A visual perturbation strategy is introduced to diversify the generated data, therefore exposing the generative model to more variant visual samples. To avoid structure disorder in the semantic space, we propose a relative positioning strategy to manipulate the attribute embeddings, guiding which to fully preserve the inter-class geometric structure. An extensive suite of experiments and analysis show that GSMFlow outperforms existing generative approaches for GZSL that suffers from the problems of the generation shifts.

\ifCLASSOPTIONcaptionsoff
  \newpage
\fi

\bibliographystyle{IEEEtran}
\bibliography{sampleBib}

\end{document}